%% file: CDC-2016.tex
\documentclass[letterpaper, 10 pt, conference]{ieeeconf}  
\IEEEoverridecommandlockouts                              
\usepackage{hyperref}
\usepackage{url}
\usepackage{bbold}
\usepackage{amsfonts}
\usepackage{amsmath,amssymb}
\usepackage{wrapfig}
\usepackage{mathrsfs} 
\usepackage{algorithm,algorithmic}
\usepackage{array}
\usepackage{times}
\usepackage{url}
\usepackage{cite}
\usepackage{upgreek}
\usepackage{float}
\usepackage{color}

\usepackage{pgfplots}
\usepackage{caption}
\usepackage{subcaption}
\usepackage{color}
\usepackage{flushend}
\usepackage{multirow}
\usepackage{rotating}
\usepackage{tikz,pgfplots}
\usepackage{standalone} 

\usepackage{theorem} 
\theoremstyle{definition}
\newtheorem{theorem}{Theorem}

\newtheorem{corollary}{Corollary}

\newtheorem{proposition}{Proposition}

\newtheorem{assumption}{Assumption}

\title{\LARGE \bf 
Online Optimization in Dynamic Environments:  Improved Regret Rates for Strongly Convex Problems
}

\author{Aryan Mokhtari, Shahin Shahrampour, Ali Jadbabaie, and Alejandro Ribeiro
\thanks{Work in this paper is supported by ARO W911NF-10-1-0388, NSF
CAREER CCF-0952867, and ONR N00014-12-1-0997. Aryan Mokhtari, Alejandro Ribeiro, and Ali Jadbabaie are with the Department of Electrical and Systems Engineering at the University of Pennsylvania, Philadelphia, PA 19104 USA. (email: {\tt aryanm, aribeiro, jadbabai@seas.upenn.edu}). Shahin Shahrampour is with the Department of Electrical Engineering at Harvard University, Cambridge, MA 02138 USA. (e-mail: {\tt shahin@seas.harvard.edu}).}
} %

\input{mysymbol.sty}

\begin{document}

\maketitle
\thispagestyle{empty}
\pagestyle{empty}

\begin{abstract}
In this paper, we address tracking of a time-varying parameter with unknown dynamics. We formalize the problem as an instance of online optimization in a dynamic setting. Using online gradient descent, we propose a method that sequentially predicts the value of the parameter and in turn suffers a loss. 
The objective is to minimize the accumulation of losses over the time horizon, a notion that is termed dynamic regret. While existing methods focus on convex loss functions, we consider strongly convex functions so as to provide better guarantees of performance. We derive a regret bound that captures the path-length of the time-varying parameter, defined in terms of the distance between its consecutive values. In other words, the bound represents the natural connection of tracking quality to the rate of change of the parameter.  We provide numerical experiments to complement our theoretical findings. 
\end{abstract}

\input{Introduction.tex}

\input{Problem-1.tex}

\input{AlgorithmRegret.tex}

\input{Simulations.tex}

\input{Conclusions.tex}

\input{Appendix.tex}

\bibliographystyle{IEEEtran}
\bibliography{IEEEabrv,shahin}

\end{document}

%% file: Introduction.tex

\section{Introduction}\label{sec_Introduction}
Convex programming is a mature discipline that has been a subject of interest among scientists for several decades \cite{boyd2004convex,nemirovskici1983problem,beck2003mirror}. The central problem of convex programming involves minimization of a convex cost function over a convex feasible set. Traditional optimization has focused on the case that the cost function is time-invariant. However, in wide range of applications, the cost function {\it i}) varies over time, and {\it ii}) there is no prior information about the dynamics of the cost function. It is therefore important to develop {\it online} convex optimization techniques, which are adapted for non-stationary environments. The problem is ubiquitous in various domains such as machine learning, control theory, industrial engineering, and operations research. 

Online optimization (learning) has been extensively studied in the literature of machine learning \cite{cesa2006prediction,shalev2011online}, proving to be a powerful tool to model sequential decisions. The problem can be viewed as a game between a learner and an adversary. The learner (algorithm) sequentially selects actions, and the adversary reveals the corresponding convex losses to the learner. The term {\it online} captures the fact that the learner receives a streaming data sequence, and it processes that adaptively.  The popular performance metric for online algorithms is called {\it regret}. Regret often measures the performance of algorithm versus a static benchmark \cite{zinkevich2003online,cesa2006prediction,freund1997decision,shalev2011online}. For instance, the benchmark could be the optimal point of the temporal average of losses, had the learner known all the losses in advance. In a broad sense, when the benchmark is a fixed sequence, the regret is called {\it static}. Furthermore, improved regret bounds are derived for the case that the losses are {\it strongly} convex \cite{hazan2007logarithmic}. 

Recent works on online learning have investigated an important direction, which involves the notion of {\it dynamic} regret \cite{zinkevich2003online,besbes2015non,hall2015online,jadbabaie2015online}. The {\it dynamic} regret can be recognized in the form of the cumulative difference between the instantaneous loss and the minimum loss. Previous works on dynamic setting investigated convex loss functions. Motivated by the fact that in the static setting curvature gives advantage to the algorithm \cite{hazan2007logarithmic}, we aim to demonstrate that {\it strong} convexity of losses yields an improved rate in dynamic setting.

Therefore, we consider the online optimization problem with strongly convex losses in dynamic setting, where the benchmark sequence varies without following any particular dynamics. We track the sequence using online gradient descent and prove that the dynamic regret can be bounded in terms of the {\it path-length} of the sequence. The path-length is defined in terms of the distance between consecutive values of the sequence. Interestingly, our result exhibits a smooth interpolation between the static and dynamic setting. In other words, the bound directly connects the tracking quality to the rate of change of the sequence. We further provide numerical experiments that verify our theoretical result.  

Of particular relevance to our setup is Kalman filtering \cite{kalman1960new}. In the original Kalman filter, there are strong assumptions on the dynamical model, such as linear state-space and Gaussian noise model. However, we depart from the classical setting by assuming no particular dynamics for the states, and observing them only through the gradients of loss functions. Instead, we provide a worst-case guarantee which captures the trajectory of the states.  

We remark that the notion of dynamic regret is also related to adaptive, shifting, and tracking regret (see e.g. \cite{herbster1998tracking,hazan2007adaptive,cesa2012new,daniely2015strongly,luo2015achieving}) in the sense of including dynamics in the problem. However, each case represents a different notion of regret. 


%% file: Problem-1.tex


\begin{table*}[]\centering\hfill
\renewcommand{\arraystretch}{1.7}
\caption{Summary of related works on dynamic online learning}
\label{tab1}
\begin{tabular}{llll}
\cline{1-4}
Reference & Regret notion & Loss function &   Regret rate
\\
\cline{1-4}
\cite{zinkevich2003online}       &$\sum_{t=1}^Tf_t(\bbx_t)-f_t(\bbu_t)$               &   Convex  & $\mathcal{O}\left(\sqrt{T}(1+{C}_T(\bbu_1,\ldots,\bbu_T))\right)$ \\ 
\cite{hall2015online}                & $\sum_{t=1}^Tf_t(\bbx_t)-f_t(\bbu_t)$              &  Convex & $\mathcal{O}\left(\sqrt{T}(1+{C}'_T(\bbu_1,\ldots,\bbu_T))\right)$\\ 
\cite{besbes2015non}              &$\sum_{t=1}^T\mathbb{E}\left[f_t(\bbx_t)\right]- f_t(\bbx^\ast_t)$  \qquad   & Convex&  $\mathcal{O}\Big(T^{2/3}{(1+V_T)}^{1/3}\Big)$ \\
\cite{besbes2015non}              &$\sum_{t=1}^T\mathbb{E}\left[f_t(\bbx_t)\right]- f_t(\bbx^\ast_t)$  \qquad   & Strongly convex&  $\mathcal{O}\Big(\sqrt{T{(1+V_T)}}\Big)$ \\
\cite{jadbabaie2015online}      &$\sum_{t=1}^Tf_t(\bbx_t)- f_t(\bbx^\ast_t)$        &  Convex   &  ${\mathcal{O}}\Big(\sqrt{D_T+1}+ \min\Big\{\sqrt{({D}_{T}+1) {C}_{T}} ,  [({D}_{T}+1)V_{T} T]^{1/3}\Big\}\Big)$\\
This work \qquad                                &$\sum_{t=1}^Tf_t(\bbx_t)- f_t(\bbx^\ast_t)$         &  Strongly convex\qquad & $\mathcal{O}\Big(1+{{C}_T}\Big)$ \\
\cline{1-4}
\end{tabular}
\end{table*}

%
\section{Problem Formulation}

We consider an online optimization problem where at each step $t$, a learner selects an action $\bbx_t\in\ccalX$ and an adversary chooses a loss function $f_t: \ccalX \to \reals$. The loss associated with the action $\bbx_t$ and function $f_t$ is given by $f_t(\bbx_t)$. 
Once the action is chosen, the algorithm (learner) receives the gradient $\nabla f_{t}(\bbx_t)$ of the loss at point $\bbx_t$.

In \textit{static} online learning, we measure the performance of the algorithm with respect to a fixed reference $\bbx \in\ccalX$ in hindsight. In other words, we aim to minimize a \textit{static regret} defined as 
\begin{equation}\label{static_regret}
{\bf Reg}_T^s(\bbx):= \sum_{t=1}^T f_t(\bbx_t) - \sum_{t=1}^T f_t(\bbx),
\end{equation}
and a particularly interesting value for $\bbx$ is $\bbx^\ast:=\argmin_{\bbx'\in \ccalX}{\sum_{t=1}^T f_t(\bbx')}$, i.e., the minimizer of the aggregate loss $\sum_{t=1}^T f_t$. A successful algorithm generates a set of actions $\{\bbx_t\}_{t=1}^T$ that yields to a sub-linear regret. Though appealing in various applications, static regret does not always serve as a comprehensive performance metric. For instance, static regret can be used in the context of static parameter estimation. However, when the parameter varies over time, we need to bring forward a new notion of regret. 

In this work, we are interested to evaluate the algorithm with respect to a more stringent benchmark, which is the sequence of instantaneous minimizers. In particular, let $\bbx_t^*:=\argmin_{\bbx\in\ccalX}f_t(\bbx)$ be the minimizer of the loss $f_t$ associated with time $t$. Then, \textit{dynamic} regret is defined as
\begin{align}\label{dynamic_regret}
{\bf Reg}_T^d(\bbx^\ast_1,\dots,\bbx^\ast_T):= \sum_{t=1}^T f_t(\bbx_t) -\sum_{t=1}^T f_t(\bbx_t^*).
\end{align}
The dynamic regret in \eqref{dynamic_regret} captures how well the action $\bbx_t$ matches the optimal action $\bbx_t^*$ for each time $t$.  It is well-known that in the worst-case, it is not possible to achieve a sub-linear dynamic regret, because drastic fluctuations in the minimum points can make the problem intractable. In this work, we would like to present a regret bound that maps the hardness of problem to variation intensity. We introduce a few complexity measures relevant to this context in the following section.

\subsection{Measures of variation and bounds on dynamic regret}
There are three common complexity measures to capture variations in the choices of the adversary. The first measure is the variation in losses $V_T$ which is characterized by
\begin{equation}\label{eqn_vt}
V_T:=\sum_{t=1}^T\sup_{\bbx \in \ccalX} \left| f_t(\bbx)-f_{t-1}(\bbx)\right|.
\end{equation}
The variation in losses $V_T$ accumulates the maximum variation between the two consecutive functions $f_t$ and $f_{t-1}$ for any feasible point $\bbx\in \ccalX$. 

The second measure of interest in dynamic settings is the variation in gradients $D_T$ which is measured by
\begin{equation}\label{eqn_dt}
D_T:=\sum_{t=1}^T \left\| \nabla f_t(\bbx_t)-M_t\right\|^2,
\end{equation}
where $M_t$ is a causally predictable sequence available to the algorithm prior to time $t$ \cite{rakhlin2013online,rakhlin2013optimization}. A simple choice is to select the previous gradient {$M_t = \nabla f_{t-1}(\bbx_{t-1})$} \cite{chiang2012online}, but $M_t$ represents any predicted value for the next objective function gradient $\nabla f_t$.

The third common measure to capture dynamics is the variation in the sequence of reference points $\bbu_{1},\dots,\bbu_T$. This variation is defined as the accumulation of the norm of the difference between subsequent  reference points
\begin{equation}\label{eqn_ct}
   C_T(\bbu_1,\dots,\bbu_T):= \sum_{t=2}^T\|\bbu_t-\bbu_{t-1}\|,
\end{equation}
The measure in \eqref{eqn_ct} accumulates variations between two arbitrary consecutive reference points $\bbu_t$ and $\bbu_{t-1}$. Whenever the reference points are the optimal points in \eqref{dynamic_regret}, i.e., whenever $\bbu_t=\bbx_t^*$, we drop the arguments in $C_T(\bbx_1^*,\dots,\bbx_T^*)$ and simply use $C_T$ to represent such variation. The variation $C_T$ captures the difference between the optimal arguments of the two consecutive losses $f_t$ and $f_{t-1}$ over time. Another notion for the variation in reference points is defined as 
 \begin{equation}\label{eqn_ctp}
C'_T(\bbu_1,\dots,\bbu_T):= \sum_{t=2}^T\|\bbu_t-\Phi_{t}(\bbu_{t-1})\|.
 \end{equation}
 where $\Phi_{t}(\bbu_{t-1})$ is the predicted reference point for step $t$ evaluated at step $t-1$ by the learner \cite{hall2015online}. If the prediction is $\Phi_{t}(\bbu_{t-1})=\bbu_{t-1}$ we recover the measure in \eqref{eqn_ct}. In general, the variation $C'_T(\bbu_1,\dots,\bbu_T)$ presents the variation of reference points with respects to a given dynamic $\Phi_{t}(\cdot)$. 

The measures in \eqref{eqn_vt}-\eqref{eqn_ctp} are different but largely compatible. They differ in that the comparisons are between functions in \eqref{eqn_vt}, gradients in \eqref{eqn_dt} and a sequence of given -- interesting is some sense, e.g., optimal -- arguments in \eqref{eqn_ct} and \eqref{eqn_ctp}, but all of them yield qualitatively comparable verdicts. The comparisons in \eqref{eqn_dt} and \eqref{eqn_ctp} further allow for the incorporation of a prediction if a model for the evolution of dynamics over time is available. The variation measures in \eqref{eqn_vt}-\eqref{eqn_ctp} have been used to bound regret in different settings with results that we summarize in Table \ref{tab1}. The work in \cite{zinkevich2003online} uses online gradient descent (OGD) with a diminishing steprsize to establish a regret of order $\mathcal{O}(\sqrt{T}(1+C_T))$ when the losses are convex. In \cite{hall2015online}, the authors study the performance of mirror descent in the dynamic setting and establish a regret bound of order $\mathcal{O}(\sqrt{T}(1+C'_T))$ when the environment follows a adynamical model $\Phi_t(\cdot)$ and the loss functions are convex. The work in \cite{besbes2015non} evaluates the performance of OGD for the case when a noisy estimate of the gradient is available and an upper bound on $V_T$ is assumed as prior knowledge. They establish regret bounds of order $\mathcal{O}(T^{2/3}(1+V_T)^{1/3})$ for convex loss functions and of order $\mathcal{O}(\sqrt{T(1+V_T)})$ for strongly convex loss functions. In \cite{jadbabaie2015online}, using optimistic mirror descent, the authors propose an adaptive algorithm which achieves a regret bound in terms of $C_T$, $D_T$, and $V_T$ simultaneously, while they assume that the learner receives each variation measure online.

Motivated by the fact that in {\it static} regret problems {\it strong} convexity results in better regret bounds -- order $\mathcal{O}(\log T)$ instead of order $\mathcal{O}(\sqrt{T})$, \cite{hazan2007logarithmic} -- we study {\it dynamic} regret problems under strong convexity assumptions. Our contribution is to show that the regret associated with the OGD algorithm (defined in Section \ref{sec:OGD}, analyzed in Section \ref{sec:analysis}) grows not faster than $1+C_T$, 
\begin{align}\label{eqn_ogd_regret_order}
   {\bf Reg}_T^d(\bbx^\ast_1,\dots,\bbx^\ast_T)
       \leq \mathcal{O}\left( 1+C_T\right) .
\end{align}
This result improves the regret bound $\mathcal{O}(\sqrt{T}(1+C_T))$ for OGD when the functions $f_t$ are convex but not necessarily strongly convex \cite{zinkevich2003online}. We remark that our algorithm assumes neither a prior knowledge nor an online feedback about $C_T$ and the only available information for the learner is the loss function gradient $\nabla f_t(\bbx_t)$.

%% file: AlgorithmRegret.tex

%
\section{Online Gradient Descent}\label{sec:OGD}  
Consider the online learning problem for $T$ iterations. At the beginning of each iteration $t$, the learner chooses the action $\bbx_t\in\ccalX$ where $\ccalX$ is a given convex set. Then, the adversary chooses a function $f_t$ and evaluates the loss associated with the iterate $\bbx_t$ which is given by the difference $f_t(\bbx_t)-f_t(\bbx_t^*)$ where $\bbx_t^*$ is the minimizer of the function $f_t$ over the set $\ccalX$. The learner does not receive the loss $f_t(\bbx_t)-f_t(\bbx_t^*)$ associated with the action $\bbx_t$. Rather, she receives the gradient $\nabla f_t(\bbx_t)$ of the cost function $f_t$ computed at $\bbx_t$. After receiving the gradient $\nabla f_t(\bbx_t)$, she uses this information to update the current iterate $\bbx_t$.

We consider a setting in which the learner uses the online gradient descent (OGD) method with a constant stepsize to update the iterate $\bbx_t$ using the released instantaneous gradient $\nabla f_t(\bbx_t)$. To be more precise, consider $\bbx_t$ as the sequence of actions that the learner chooses and define $\hbx_t\in \ccalX$ as a sequence of auxiliary iterates. At each iteration $t$, given the iterate $\bbx_t$ and the instantaneous gradient $\nabla f_t(\bbx_t)$, the learner computes the auxiliary variable $\hbx_t$ as
\begin{equation}\label{auxiliary_update}
\hbx_t= \Pi_\ccalX\left(\bbx_t-\frac{1}{\gamma} \nabla f_t(\bbx_t)\right),
\end{equation}
where $\gamma$ is a positive constant and $\Pi_\ccalX$ denotes the projection onto the nearest point in the set $\ccalX$, i.e, $\Pi_\ccalX(\bby)=\argmin_{\bbx\in\ccalX}\|\bbx-\bby\|$. Then, the action $\bbx_{t+1}$ is evaluated as 
\begin{equation}\label{variable_update}
\bbx_{t+1}= \bbx_t+h(\hbx_t-\bbx_t),
\end{equation}
where $h$ is chosen from the interval~$(0,1]$. The online gradient descent method is summarized in Algorithm \ref{algo_OGD}.

The updated action $\bbx_{t+1}$ in \eqref{variable_update} can be written as $\bbx_{t+1}= (1-h)\bbx_t+h\hbx_t$. Therefore, we can reinterpret the updated action $\bbx_{t+1}$ as a weighted average of the previous iterate $\bbx_t$ and the auxiliary variable $\hbx_t$ which is evaluated using the gradient of the function $f_t$. It is worth mentioning that for a small choice of $h\approx 0$, $\bbx_{t+1}$ is close to the previous action $\bbx_t$, while the previous action $\bbx_t$ has less impact on $\bbx_{t+1}$ when $h$ is close to 1. 

The auxiliary variable $\hbx_t$ is the result of applying projected gradient descent on the current iterate $\bbx_t$ as shown in \eqref{auxiliary_update}. This update can be interpreted as minimizing a first-order approximation of the cost function $f_t$ added to a proximal term $(\gamma/2)\|\bbx-\bbx_t\|^2$ as we show in the following proposition. 

%
\begin{algorithm}[t]
\caption{Online Gradient Descent}\label{algo_OGD} 
\begin{algorithmic}[1] 
\small{\REQUIRE Initial vector $\bbx_{1}\in \cal{X}$, constants $h$ and $ \gamma$.
\FOR {$t=1,2,\ldots, T$}
   \STATE Play $\bbx_t$
   \STATE Observe the gradient of the current action $\nabla f_{t}(\bbx_t)$
   \STATE Compute the auxiliary var.: 
   	 $ \displaystyle{\quad \hbx_t= \Pi_\ccalX\left(\bbx_t-\frac{1}{\gamma} \nabla f_t(\bbx_t)\right)}$
   \STATE Compute the next action: 
          $ \displaystyle{\quad \bbx_{t+1} =\bbx_t+h(\hbx_t-\bbx_t)}$
\ENDFOR}
\end{algorithmic}\end{algorithm}

\begin{proposition}\label{prop_update}
\vspace{-1mm}
Consider the update in \eqref{auxiliary_update}. Given the iterate $\bbx_t$, the instantaneous gradient $\nabla f_{t}(\bbx_t)$, and the positive constant $\gamma$, the optimal argument of the optimization problem 
\begin{align}\label{eqi_update}
\tbx_t
                 =  \argmin_{\bbx\in \cal{X}} 
                 \left\{ \nabla f_{t}(\bbx_t)^T(\bbx-\bbx_t) 
                 	+\frac{\gamma}{2} \|\bbx-\bbx_t \|^2 \right\},
\end{align}
is equal to the iterate $\hbx_t$ generated by \eqref{auxiliary_update}.

\end{proposition} 
\begin{myproof} 
\vspace{-1mm}
See Section \ref{prop_update_app}. \end{myproof}

The result in Proposition \ref{prop_update} shows that the updates in \eqref{auxiliary_update} and \eqref{eqi_update} are equivalent. In the implementation of OGD we use the update in \eqref{auxiliary_update}, since the computational complexity of the update in \eqref{auxiliary_update} is lower than the complexity of the minimization in \eqref{eqi_update}. On the other hand, the update in \eqref{eqi_update} is useful in the regret analysis of OGD that we undertake in the following section.

%
\section{Regret Analysis }\label{sec:analysis}

We proceed to show that the dynamic regret ${\bf Reg}_T^d(\bbx^\ast_1,\dots,\bbx^\ast_T)={\bf Reg}_T^d$ defined in \eqref{dynamic_regret} associated with the actions $\bbx_t$ generated by the online gradient descent algorithm in \eqref{variable_update} has an upper bound on the order of the variation in the sequence of optimal arguments $C_T=\sum_{t=2}^T\|\bbx_{t}^*-\bbx_{t-1}^*\|$. In proving this result, we assume the following conditions are satisfied.

\begin{assumption}\label{strong_convexity_ass}
The functions $f_t$ are strongly convex over the convex set $\cal{X}$ with constant $\mu>0$, i.e., 
\begin{equation}\label{strong_convex}
f_t(\bbx)\geq f_t(\bby) +\nabla f_t(\bby)^T(\bbx-\bby) + \frac{\mu}{2} \|\bbx-\bby\|^2,
\end{equation}
for any $\bbx,\bby\in \cal{X}$ and $1\leq t\leq T$.
\end{assumption}

\begin{assumption}\label{Lipschitz_gradients_ass}
The gradients $\nabla f_t$ are Lipschitz continuous    over the set $\cal{X}$ with constant $L<\infty$, i.e., 
\begin{equation}\label{lip_cont}
\|\nabla f_t(\bby)-\nabla f_t(\bbx)\| \leq L \|\bbx-\bby\|,
\end{equation}
for any $\bbx,\bby\in \cal{X}$ and $1\leq t\leq T$.
\end{assumption}

\begin{assumption}\label{bounded_grad_ass}
The gradient norm $\|\nabla f_t\|$ is bounded above by a positive constant $G$ or equivalently 
\begin{equation}\label{bounded_gradient}
\sup_{\bbx\in\ccalX,1\leq t\leq T}\|\nabla f_{t}(\bbx)\|\leq G.
\end{equation}
\end{assumption}

According to Assumption \ref{strong_convexity_ass}, the instantaneous functions $f_t$ are strongly convex over the convex set $\ccalX$ which  implies that there exists a unique minimizer $\bbx_t^*$ for the function $f_t$ over the convex set $\ccalX$. The Lipschitz continuity of the gradients $\nabla f_t$ in Assumption \ref{Lipschitz_gradients_ass} is customary in the analysis of descent methods. Notice that we only assume for a fixed function $f_t$ the gradients are Lipschitz continuous  and we do not assume any conditions on the difference of two gradients associated with two different instantaneous functions. To be more precise, there is no condition on the norm $\|\nabla f_t(\bby)-\nabla f_{t'}(\bbx)\|$ where $t\neq t'$. The bound on the gradients norm in Assumption \ref{bounded_grad_ass} is typical in the analysis of online algorithms for constrained optimization.

Our main result on the regret bound of OGD in dynamic settings is derived from the following proposition that bounds the difference $\|\bbx_{t+1}-\bbx_t^*\| $ in terms of the distance $\|\bbx_{t}-\bbx_t^*\|$.

\begin{proposition}\label{prop_lin}
Consider the online gradient descent method (OGD) defined by \eqref{auxiliary_update} and \eqref{variable_update} or the equivalent \eqref{eqi_update}. Recall the definition of $\bbx_t^*$ as the unique minimizer of the function $f_t$ over the convex set $\cal{X}$. If Assumptions \ref{strong_convexity_ass} and \ref{Lipschitz_gradients_ass} hold and the stepsize parameter $\gamma$ in \eqref{auxiliary_update} is chosen such that $\gamma\geq L$, then the sequence of actions $\bbx_t$ generated by OGD satisfies 
\begin{equation}\label{lin_convg_claim}
\|\bbx_{t+1}-\bbx_t^*\| \leq \rho\ \! \|\bbx_{t}-\bbx_t^*\|,
\end{equation}
where $ 0 \leq \rho:=(1-h\mu/\gamma)^{1/2} < 1$ is a non-negative constant strictly smaller than $1$.

\end{proposition} 
\begin{myproof}See Section \ref{prop_lin_app}.\end{myproof}

The result in Proposition \ref{prop_lin} shows that the distance between the action $\bbx_{t+1}$ and the optimal argument $\bbx_t^*$ is strictly smaller than the difference between the previous action $\bbx_{t}$ and the optimal argument $\bbx_t^*$ at step $t$. The inequality in \eqref{lin_convg_claim} implies that if the optimal arguments of the functions $f_t$ and $f_{t+1}$ which are $\bbx_t^*$ and $\bbx_{t+1}^*$, respectively, are not far away from each other the iterates $\bbx_t$ can track the optimal solution sequence $\bbx_t^*$. Notice that if in the left hand side of \eqref{lin_convg_claim} instead of $\bbx_t^*$ we had the optimal argument $\bbx_{t+1}^*$ at step $t+1$, then we could show that the sequence of actions $\bbx_t$ generated by OGD asymptotically converges to the sequence of optimal arguments $\bbx_t^*$. Thus, the performance of OGD depends on the rate that the sequence of optimal arguments changes. This conclusion is formalized in the following Theorem.

\begin{theorem}\label{thm_OGD} Consider the online gradient descent method (OGD) defined by \eqref{auxiliary_update} and \eqref{variable_update} or the equivalent \eqref{eqi_update}.  Suppose that the constant $h$ is chosen from the interval $(0,1]$ and the constant $\gamma$ satisfies the condition $\gamma\geq L$, where $L$ is the gradients Lipschitz continuity constant. If Assumptions \ref{strong_convexity_ass} and \ref{Lipschitz_gradients_ass} hold, then the sequence of actions $\bbx_t$ generated by OGD satisfies 
\begin{equation}\label{lin_convg_claim2}
\sum_{t=1}^T \|\bbx_{t}-\bbx_t^*\|\leq K_1 \sum_{t=2}^T\|\bbx_{t}^*-\bbx_{t-1}^*\|+K_2,
\end{equation}
where the constants $K_1$ and $K_2$ are explicitly given by
\begin{align}\label{C_defs}
K_1:= \frac{\|\bbx_{1}-\bbx_1^*\| -\rho \| \bbx_T-\bbx_T^*\| }{(1-\rho)}, \quad K_2:= \frac{1}{(1-\rho)}. 
\end{align}
\end{theorem} 

\begin{myproof} See Section \ref{thm_OGD_app}. \end{myproof}

From Theorem \ref{thm_OGD}, we obtain an upper bound for the aggregate variable error $\sum_{t=1}^T \|\bbx_{t}-\bbx_t^*\|$ in terms of the aggregate variation in the optimal arguments $C_T=\sum_{t=2}^T\|\bbx_{t}^*-\bbx_{t-1}^*\|$. This result matches the intuition that for the scenarios that the sequence of the optimal arguments $\{\bbx_t^*\}_{t=1}^T$ is not varying fast, the sequence of actions generated by the online gradient descent method can achieve a sublinear regret bound. In particular, when the optimal arguments are all equal to each other, i.e., when $\bbx_1^*=\dots=\bbx_T^*$, the aggregate error $\sum_{t=1}^T \|\bbx_{t}-\bbx_t^*\|$ is bounded above by the constant $K_2$ which is independent of $T$. We use the bounded gradients assumption in \eqref{bounded_gradient} to translate the result in \eqref{lin_convg_claim2} into an upper bound for the dynamic regret ${\bf Reg}_T^d$ defined in \eqref{dynamic_regret}.

\begin{corollary} \label{final_result}
Adopt the same definitions and hypothesis of Theorem \ref{thm_OGD} and furhter assume that the gradient norms $\|\nabla f_t(\bbx)\|$ are upper bounded by the constant $G$ for all $\bbx\in\ccalX$ as in Assumption \ref{bounded_grad_ass}. Then, the dynamic regret ${\bf Reg}_T^d$ for the sequence of actions $\bbx_t$ generated by OGD is bounded above by
\begin{equation}\label{col_claim}
{\bf Reg}_T^d \leq G K_1 \sum_{t=2}^T\|\bbx_{t}^*-\bbx_{t-1}^*\|+G K_2.
\end{equation}
\end{corollary} 
\begin{myproof}
Notice that for any vectors $\bbx$ and $\bby$ we know that there exists a vector $\bbz$ such that  
$f_t(\bbx)-f_t(\bby)= \nabla f_t(\bbz)^T(\bbx-\bby)$ where $\bbz$ is a point from the set $\{\bbv\mid \bbv = \alpha \bbx +(1-\alpha)\bby,\ 0\leq\alpha\leq1 \}$. Thus, the objective function difference $|f_t(\bbx)-f_t(\bby)|$ is bounded above by $G\|\bbx-\bby\|$. Setting $\bbx=\bbx_t$ and $\bby=\bbx_t^*$ implies that 
\begin{equation}\label{sososo}
f_t(\bbx_t)-f_t(\bbx_t^*) \leq G \|\bbx_t-\bbx_t^*\|. 
\end{equation}
By summing up \eqref{sososo} for all times $t$, we conclude that the dynamic regret is bounded above as ${\bf Reg}_T^d=\sum_{t=1}^T f_t(\bbx_t)-f_t(\bbx_t^*)\leq G \sum_{t=1}^T \|\bbx_t-\bbx_t^*\|$. This observation combined with the result in \eqref{lin_convg_claim2} yields \eqref{col_claim}.\end{myproof}

%
The result in Corollary \ref{final_result} states that under the conditions that the functions $f_t$ are strongly convex and their gradients are bounded and Lipschitz continuous, the dynamic regret ${\bf Reg}_T^d$ associated with the online gradient descent method satisfies the order bound that we previewed in \eqref{eqn_ogd_regret_order}. As already mentioned, this bound improves the OGD rate $\mathcal{O}(\sqrt{T}(1+C_T))$ when the functions $f_t$ are convex but not ncessarily strongly convex and the stepsize is diminishing \cite{zinkevich2003online}. 

Some interesting conclusions can be derived if we consider specific rates of variability: 

\medskip\noindent{\bf Constant functions. } If the functions are constant, i.e., if $f_t=f$ for all times, we have $C_T=0$ and it follows that the regret grows at a rate $\mathcal{O}(1)$. This means that $\bbx_t$ converges to $\bbx^*$ and we recover a convergence proof for gradient descent.

\medskip\noindent{\bf Linearly decreasing variability.} If the difference between consecutive arguments decreases as $1/t$, we have that $C_T = \mathcal{O}(\log T)$ and that the regret then grows at a logarithmic rate as well. Since this implies that the normalized regret grows not faster than ${\bf Reg}_T^d(\bbx^\ast_1,\dots,\bbx^\ast_T)/T \leq \mathcal{O}(\log T/T)$ we must have that $\bbx_t$ converges to $\bbx_t^*$.

\medskip\noindent{\bf Decreasing variability.} If the variability decreases as $1/t^{\alpha}$ with $\alpha\in(0,1)$, the regret is of order $\mathcal{O}(1+C_T) = \mathcal{O}(1 + T^{1-\alpha})$. As before, this must imply that $\bbx_t$ converges to $\bbx_t^*$.

\medskip\noindent{\bf Constant variability.} If the variability between functions stays constant, say $\|\bbu_t^*-\bbu_{t-1}^*\|\leq C$ for all $T$, the regret is of order $\mathcal{O}(1+C_T) = \mathcal{O}(1 + CT)$. The normalized regret is then of order ${\bf Reg}_T^d(\bbx^\ast_1,\dots,\bbx^\ast_T)/T \leq \mathcal{O}(C)$. This means that we have a steady state tracking error, where the tracking error depends on how different adjacent functions are.

%% file: Simulations.tex

%
\section{Numerical Experiments}\label{sec:simulations}
 
We numerically study the performance of OGD in solving a sequence of quadratic programming problems. Consider the decision variable $\bbx=[x_1;x_2]\in\reals^2$ and the quadratic function $f_t$ at time $t$ which is defined as
\begin{align}\label{sim_loss}
f_t(\bbx)\!=f_t(x_1,x_2)\!=\rho\|x_1-a_t\|^2+\|x_2-b_t\|^2+c_t,
\end{align}
where $a_t,b_t,$ and $c_t$ are time-variant scalars and $\rho>0$ is a positive constant. The coefficient $\rho$ controls the condition number of the objective function $f_t$. In particular for $\rho>1$, the problem condition number is equal to $\rho$. The convex set $\ccalX$ is defined as $x_1^2+x_2^2=r^2$ which is the circle with center $[0;0]$ and radius $r$. The radius $r$ is chosen such that the optimal argument of the function $f_t$ over $\reals^2$, which is $[a_t,b_t]$, is not included in the set $\ccalX$. This way we ensure that the constraint $\bbx\in\ccalX$ is active at the optimal solution.

In our experiments we pick $\rho=100$ to have a quadratic optimization problem with large condition number $100$. Note that if we choose $\rho=1$, the condition number of the function $f_t$ is $1$ and OGD can minimize the cost $f_t$ in a couple of iterations. The constant $\gamma$ in OGD is set as $\gamma=2\rho$ in all experiments, since the Lipschitz continuity constant of gradients is $L=2\rho$. Moreover, the OGD parameter $h$ is set as $h=1$ which implies $\bbx_{t+1}=\hbx_t$.

\begin{figure*}[ht]
   \centering
\begin{subfigure}[b]{0.32\textwidth}
\includegraphics[width=\linewidth]{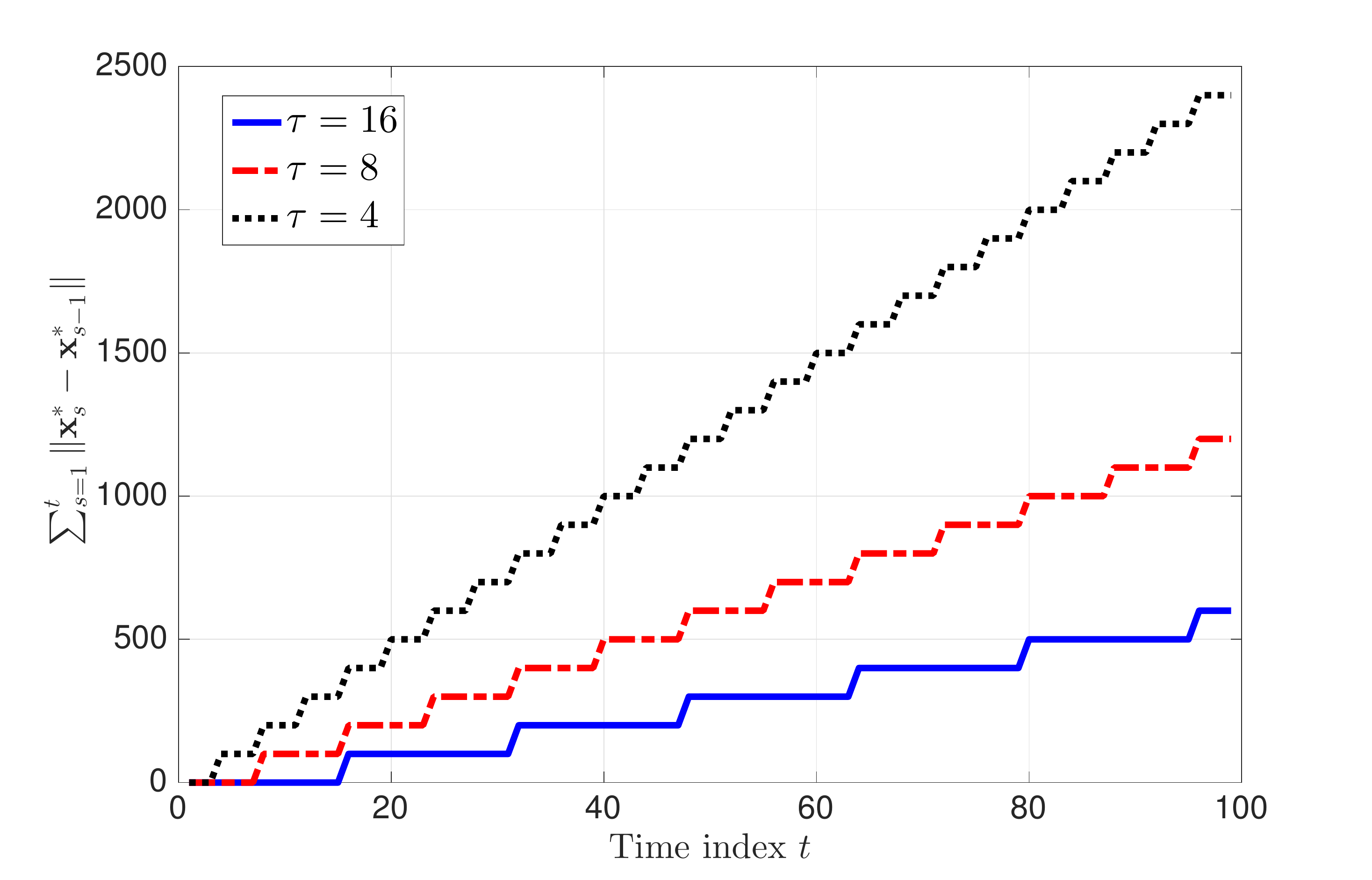}
\caption{}
\label{subfig:C}
\end{subfigure}
~
\begin{subfigure}[b]{0.32\textwidth}
\includegraphics[width=\linewidth]{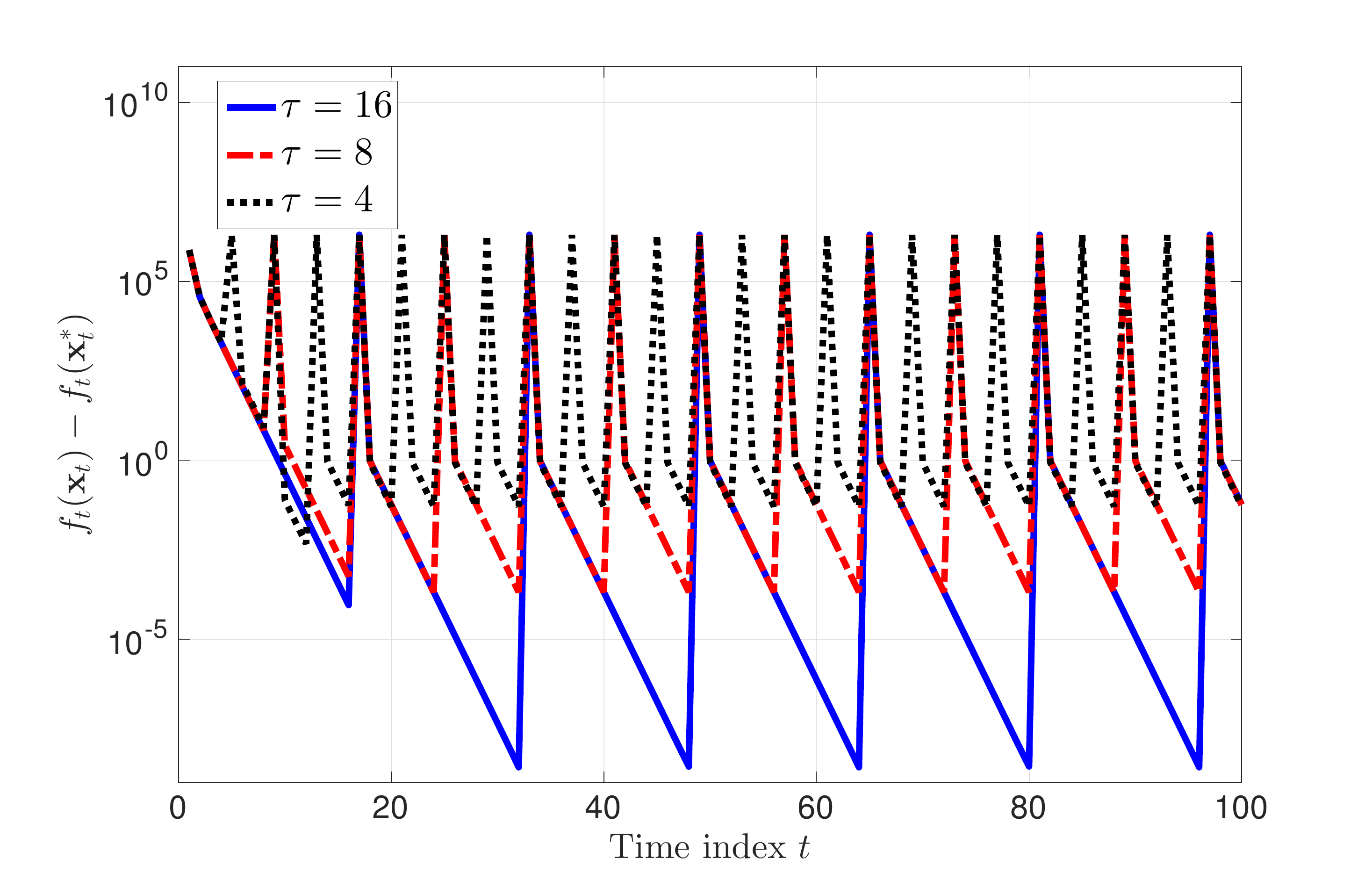}
\caption{}
\label{subfig:obj}
\end{subfigure}
~
\begin{subfigure}[b]{0.32\textwidth}
\includegraphics[width=\linewidth]{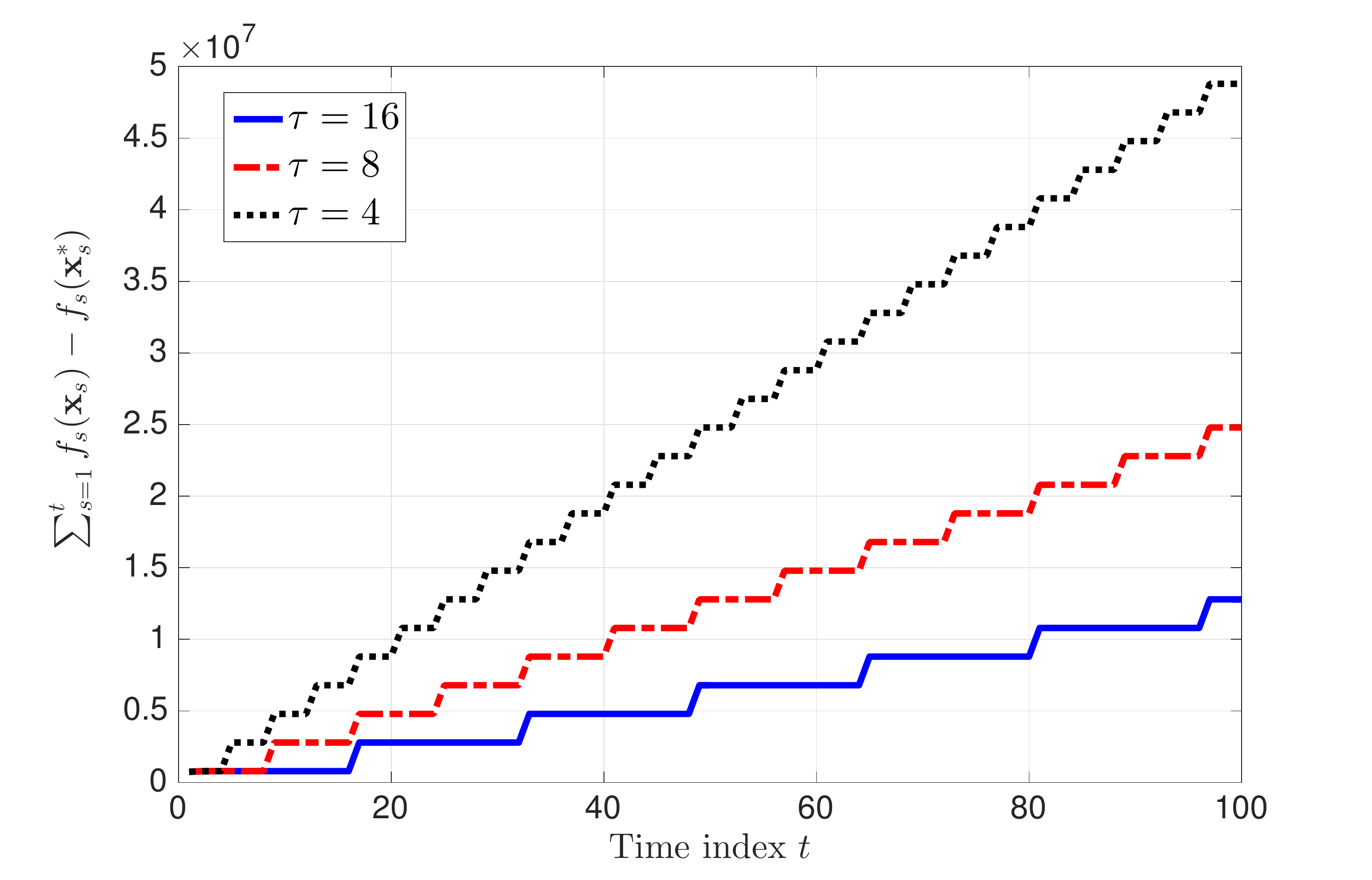}
\caption{}
\label{subfig:reg}
\end{subfigure}
\caption{Performance of OGD for the case that the adversary switches between two quadratic functions with different minimizers every $\tau$ iterations. The variation in the sequence of optimizers $C_t= \sum_{s=2}^t\|\bbx_s^*-\bbx_{s-1}^*\|$, instantaneous objective function error $ f_t(\bbx_t)-f_t(\bbx_t^*)$, and dynamic regret ${\bf Reg}_t^d= \sum_{s=1}^t f_s(\bbx_s)-f_s(\bbx_s^*)$ are shown in Figures \ref{subfig:C}, \ref{subfig:obj}, and \ref{subfig:reg}, respectively. In the case that $\tau$ is small and the adversary switches more often between the two quadratic functions, the variation $C_t$ and the dynamic regret ${\bf Reg}_t^d$ grow faster. Moreover, the growth patterns of $C_t$ and ${\bf Reg}_t^d$ are similar. } \label{fig_1}
\end{figure*}

To characterize the instantaneous performance of OGD we define $ f_t(\bbx_t)-f_t(\bbx_t^*)$ as the instantaneous objective function error at time $t$. Further, we define ${\bf Reg}_t^d :=\sum_{s=1}^t f_s(\bbx_s)-f_s(\bbx_s^*)$ as the dynamic regret up to step $t$. Likewise, we define $ C_t:= \sum_{s=2}^t\|\bbx_s^*-\bbx_{s-1}^*\|$ as the total optimal argument variation until step $t$.

We consider two different cases to study the regret bound of OGD in dynamic online settings. First, we consider a switching problem that the adversary switches between two quadratic functions after a specific number of iterations. Then, we study the case that the sequence of optimal arguments $\bbx_t^*$ changes at each iteration, while the difference $\|\bbx_t^*-\bbx_{t-1}^*\|$ diminishes as time progresses.

%

\subsection{Switching problem}

Consider the case that the adversary chooses between two functions where each of them is of the form of the quadratic function $f_t$ in \eqref{sim_loss}. In particular, consider the case that the adversary chooses the parameters $a_t$, $b_t$, and $c_t$ from the two sets $\ccalS^{(1)}=\{a,b,c\}$ and $\ccalS^{(2)}=\{a',b',c'\}$. Therefore, at each iteration the adversary chooses either $f^{(1)}=\rho\|x_1-a\|^2+\|x_2-b\|^2+c$ or $f^{(2)}=\rho\|x_1-a'\|^2+\|x_2-b'\|^2+c'$. We run OGD for a fixed number of iterations $T=100$ and assume that the adversary switches between the functions $f^{(1)}$ and $f^{(2)}$ every $\tau$ iterations. 

In our experiments we set $a=-100$, $b=0$, $c=30$, $a'=100$, $b'=20$, and $c'=-50$. The convex set $\ccalX$ is defined as $x_1^2+x_2^2=50^2$ which is the circle with center $[0;0]$ and radius $50$. Note that this circle does not contain the points $[a;b]=[-100;0]$ and $[a';b']=[100;20]$. The optimal argument of $f^{(1)}$ and $f^{(2)}$ over the convex set $\ccalX$ are  ${\bbx^{(1)}}^*=[-50;0]$ and ${\bbx^{(2)}}^*=[49.99;0.19]$, respectively. We set the initial iterate $\bbx_0=[0;40]$ which is a feasible point for the set $\ccalX$. We consider three different cases that $\tau=4$, $\tau=8$, and $\tau=16$. The performance of OGD for the three different choices of $\tau$ are illustrated in Figure \ref{fig_1}.

Figure \ref{subfig:C} demonstrates the variable variation $C_t:= \sum_{s=2}^t\|\bbx_s^*-\bbx_{s-1}^*\|$ over time $t$. For the case $\tau=16$, the value of $C_t$ increases every $16$ iterations and the increment is equal to the norm of the difference between the optimal arguments of $f^{(1)}$ and $f^{(2)}$ which is $\|{\bbx^{(1)}}^*-{\bbx^{(2)}}^*\|=100$. After $T=100$ iterations, we observe $6$ jumps which implies that the total variable variation is $C_T=600$. Likewise, for the cases that $\tau=8$ and $\tau=4$, we observe $12$ and $24$ jumps in their corresponding plots, and the aggregate variable variations are $C_T=1200$ and $C_T=2400$, respectively.

Figure \ref{subfig:obj} showcases the instantaneous function error $ f_t(\bbx_t)-f_t(\bbx_t^*)$ versus number of iterations $t$ for $\tau=16$, $\tau=8$, and $\tau=4$. In all of the cases, the sequence of errors $ f_t(\bbx_t)-f_t(\bbx_t^*)$ converges linearly to $0$ until the time the adversary switches the objective function $f_t$. By increasing the number of times that the adversary switches between the functions $f^{(1)}$ and $f^{(2)}$, the phase of linear convergence becomes shorter and the algorithm restarts more often. 

\begin{figure*}[t]
   \centering
\begin{subfigure}[b]{0.32\textwidth}
\includegraphics[width=\linewidth]{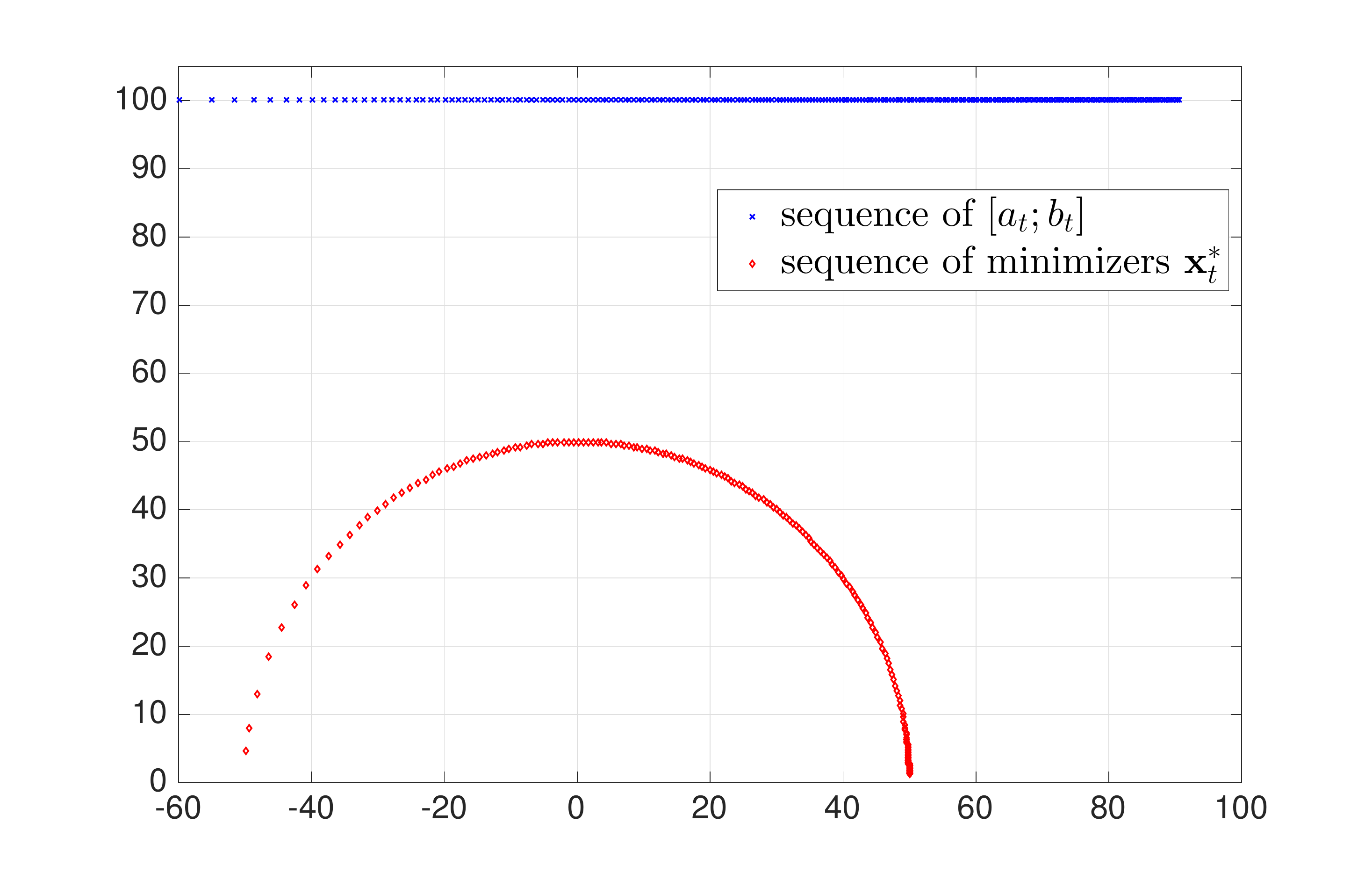}
\caption{}
\label{subfig:path}
\end{subfigure}
~
\begin{subfigure}[b]{0.32\textwidth}
\includegraphics[width=\linewidth]{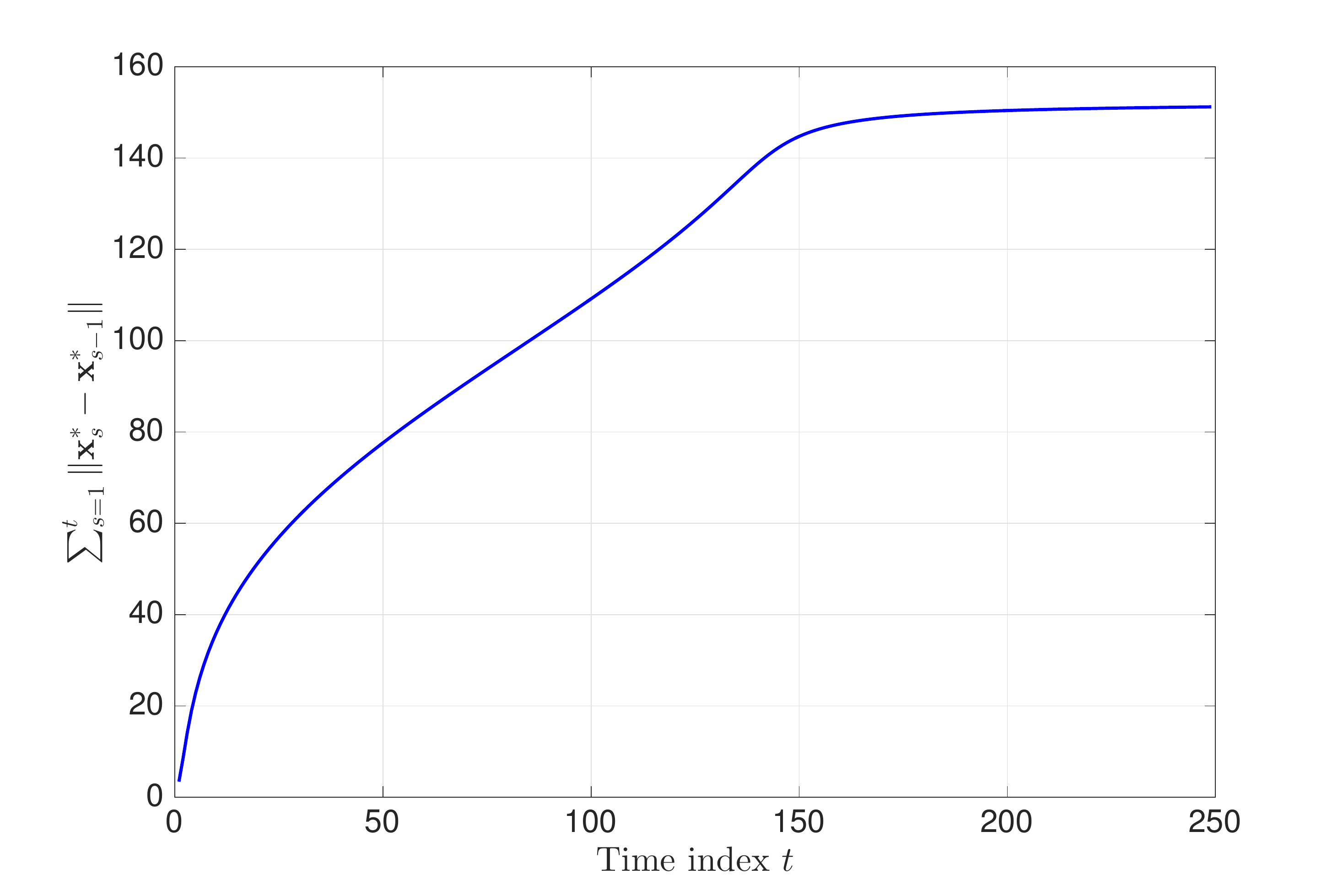}
\caption{}
\label{subfig:CTB}
\end{subfigure}
~
\begin{subfigure}[b]{0.32\textwidth}
\includegraphics[width=\linewidth]{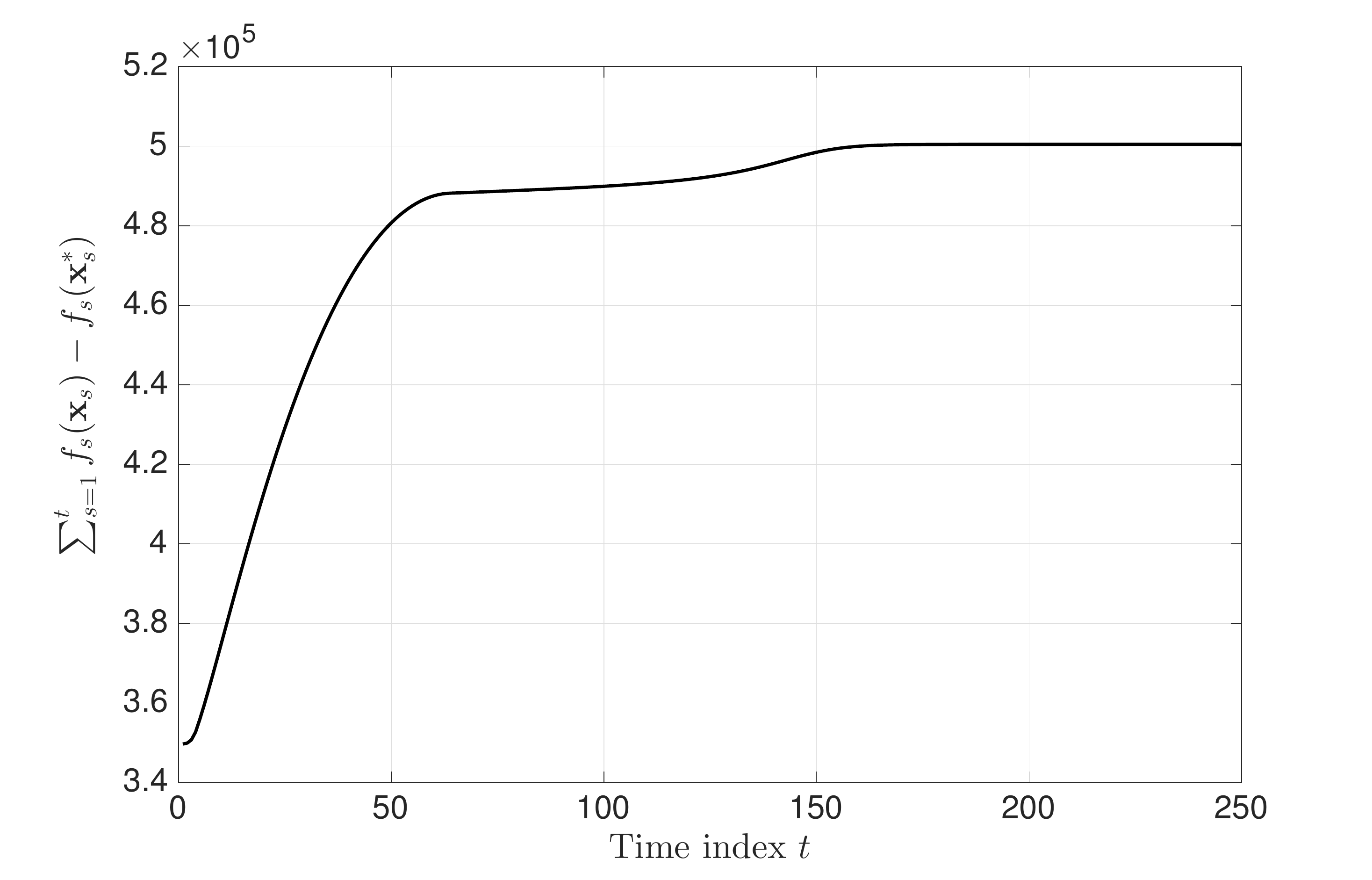}
\caption{}
\label{subfig:regB}
\end{subfigure}
\caption{Performance of OGD for the case that the adversary chooses a sequence of quadratic functions $f_t$ where the sequence of optimal arguments $\bbx_t^*$ is convergent. Figure \ref{subfig:path} illustrates the paths for the function parameters $[a_t;b_t]$ and the optimal arguments $\bbx_t^*$ over the set $\ccalX$.
The variation in the sequence of optimizers $C_t:= \sum_{s=2}^t\|\bbx_s^*-\bbx_{s-1}^*\|$ and the dynamic regret ${\bf Reg}_t^d =\sum_{s=1}^t f_s(\bbx_s)-f_s(\bbx_s^*)$ are shown in Figures \ref{subfig:CTB} and \ref{subfig:regB}, respectively. The sequence of optimal arguments is convergent in a way that the variation in the sequence of optimizers $C_t:= \sum_{s=2}^t\|\bbx_s^*-\bbx_{s-1}^*\|$ is summable. Likewise, the dynamic regret ${\bf Reg}_T^d$ associated with the OGD method does not grow as the total number of iterations $T$ increases.} \label{fig_2}
\end{figure*}

Thus, we expect to observe larger regret for the scenarios that $\tau$ is smaller. The dynamic regret ${\bf Reg}_t^d =\sum_{s=1}^t f_s(\bbx_s)-f_s(\bbx_s^*)$ versus the number of iterations $t$ is shown in Figure \ref{subfig:reg} for $\tau=16$, $\tau=8$, and $\tau=4$. As we expect, for the case that $\tau=4$ the dynamic regret ${\bf Reg}_t^d$ grows faster. Note that the dynamic regret ${\bf Reg}_T^d$ of OGD for $\tau=16$, $\tau=8$, and $\tau=4$ after $T=100$ iterations are ${\bf Reg}_T^d=1.28\times 10^7$, ${\bf Reg}_T^d=2.48\times 10^7$, and ${\bf Reg}_T^d=4.88\times 10^7$, respectively. 

Comparing the variation $C_t$ and the dynamic regret ${\bf Reg}_T^d$ for the three cases $\tau=16$, $\tau=8$, and $\tau=4$ shows that the growth patterns of $C_t$ and ${\bf Reg}_T^d$ are similar. This observation is consistent with the theoretical result in Corollary \ref{final_result} which indicates that the upper bound for the dynamic regret ${\bf Reg}_T^d$ is of order $\mathcal{O}(1+C_T)$.

%
\subsection{Diminishing variations} 

In this section we consider the case that the adversary picks a sequence of functions $f_t$ as in \eqref{sim_loss} such that the sequence of optimizers $\bbx_t^*$ is convergent. 

We use the parameters in Figure~\ref{fig_1} except the total number of iterations which is set as $T=250$. We set the initial values for $[a_1;b_1]$ as $[-60;100]$. Further, we assume that $b_t$ is time-invariant and for all steps $t$ we have $b_t=b_1=100$. On the other hand, we assume that the parameter $a_t$ changes as time passes and it satisfies the recursive formula $a_{t+1}=a_t+5 \sqrt{1/t}$. The sequence of parameters $[a_t,b_t]$, which are the optimal arguments of the function $f_t$ over $\reals^2$, is illustrated in Figure \ref{subfig:path}. Moreover, the set of optimal arguments of $f_t$ over the set $\ccalX$, which are indicated by $\bbx_t^*$, are also demonstrated in Figure \ref{subfig:path}. This plot shows that as time progresses and the difference between the functions $f_t$ and $f_{t-1}$ becomes less significant, the difference between the optimal arguments $\bbx_{t}^*$ and $\bbx_{t-1}^*$ diminishes. To formally study the variation in the sequence of optimal arguments $\bbx_t^*$, we demonstrate the variable variation $C_t:= \sum_{s=2}^t\|\bbx_s^*-\bbx_{s-1}^*\|$ in terms of number of iterations $t$ in Figure~\ref{subfig:CTB}. As we expect, the variation $C_t$ converges as time progresses, since the difference $\|\bbx_{t}^*-\bbx_{t-1}^*\|$ is diminishing.

Figure \ref{subfig:regB} illustrates the dynamic regret ${\bf Reg}_t^d =\sum_{s=1}^t f_s(\bbx_s)-f_s(\bbx_s^*)$ of OGD in terms of number of iterations $t$. We observe that the dynamic regret of OGD follows the pattern of the variation $C_t$ and converges to a constant value as time progresses. Thus, the dynamic regret ${\bf Reg}_T^d$ does not grow with the number of iterations $T$. This observation matches the theoretical result in Corollary \ref{final_result} that the dynamic regret ${\bf Reg}_T^d$ converges to a constant value when the variation in optimal arguments $C_T:= \sum_{s=2}^T\|\bbx_s^*-\bbx_{s-1}^*\|$ does not grow by the number of iterations $T$.

%% file: Conclusions.tex

\section{Conclusions}\label{sec_conclusions}

This paper studies the performance of the online gradient descent (OGD) algorithm in online dynamic settings. We established an upper bound for the dynamic regret of OGD in terms of the variation in the sequence of optimal arguments $\bbx_t^*$ defined by $C_T= \sum_{t=2}^T\|\bbx_t^*-\bbx_{t-1}^*\|$. We showed that if the functions $f_t$ chosen by the adversary are strongly convex, the online gradient descent method with a proper constant stepsize has a regret of order $\mathcal{O}(1+C_T)$. This result indicates that the dynamic regret bound of OGD for strongly convex functions is significantly smaller than the regret bound of order $\mathcal{O}(\sqrt{T}(1+C_T))$ for convex settings. Numerical experiments on a dynamic quadratic programming verified our theoretical result that the dynamic regret of OGD has an upper bound of order $\mathcal{O}(1+C_T)$.

%% file: Appendix.tex

\section{APPENDIX}

\subsection{Proof of Proposition \ref{prop_update}}\label{prop_update_app}

Let's define $\tbx_t$ as the minimizer of the program $\tbx_t
                : =  \argmin_{\bbx\in \cal{X}} 
                 \left\{ \nabla f_{t}(\bbx_t)^T(\bbx-\bbx_t) 
                 	+({\gamma}/{2}) \|\bbx-\bbx_t \|^2 \right\}$ and $\hbx^t$ as defined in \eqref{auxiliary_update}. Our goal is to show that $\hbx_t=\tbx_t$. Based on the definition of the update in \eqref{auxiliary_update} we can write 
\begin{equation}\label{proof_update_1}
\left\|\hbx_{t}\!-\!\left[\bbx_t-\frac{1}{\gamma}\nabla f_t(\bbx_t)\right]\right\|\leq \left\|\bby\!-\!\left[\bbx_t-\frac{1}{\gamma}\nabla f_t(\bbx_t)\right]\right\|,
\end{equation}
for any $\bby\in\ccalX$. Now assume that $\hbx_t\neq\tbx_t$. Therefore, since the set $\ccalX$ is convex the minimization $\min_{\bbx\in\ccalX}\|\bbx-(\bbx_t-({1}/{\gamma})\nabla f_t(\bbx_t))\|$ has a unique solution which is $\hbx_t$. Therefore, by setting $\bby=\tbx_t$ in \eqref{proof_update_1} we can replace $\leq$ by $<$ which implies that
\begin{equation}\label{proof_update_2}
\left\|\hbx_{t}\!-\!\left[\bbx_t-\frac{1}{\gamma}\nabla f_t(\bbx_t)\right]\right\|\!<\left\|\tbx_t\!-\!\left[\bbx_t-\frac{1}{\gamma}\nabla f_t(\bbx_t)\right]\right\|\!,
\end{equation}
Computing the square of both sides of \eqref{proof_update_2}, simplifying the resulted expression, and multiplying both sides by $\gamma/2$ yield
\begin{align}\label{proof_update_3}
&\frac{\gamma}{2}\left\|\hbx_{t}-\bbx_t\right\|^2
+(\hbx_{t}-\bbx_t)^T\nabla f_t(\bbx_t)\nonumber\\
&\quad 
< \frac{\gamma}{2}\left\|\tbx_{t}-\bbx_t\right\|^2
+(\tbx_{t}-\bbx_t)^T\nabla f_t(\bbx_t).
\end{align}
However, we know that $\tbx_t$ is a minimizer of the program $\min_{\bbx\in \cal{X}} 
                 \left\{ \nabla f_{t}(\bbx_t)^T(\bbx-\bbx_t) 
                 	+({\gamma}/{2}) \|\bbx-\bbx_t \|^2 \right\}$. The optimality condition of this minimization implies that 
\begin{align}\label{proof_update_4}
 &\frac{\gamma}{2}\left\|\tbx_{t}-\bbx_t\right\|^2
+(\tbx_{t}-\bbx_t)^T\nabla f_t(\bbx_t)\nonumber\\
&\quad 
\leq 
 \frac{\gamma}{2}\left\|\bby-\bbx_t\right\|^2
+(\bby-\bbx_t)^T\nabla f_t(\bbx_t),
\end{align}
for any $\bby\in\ccalX$. By setting $\bby=\hbx_t$ in \eqref{proof_update_4} we obtain that
\begin{align}\label{proof_update_5}
 &\frac{\gamma}{2}\left\|\tbx_{t}-\bbx_t\right\|^2
+(\tbx_{t}-\bbx_t)^T\nabla f_t(\bbx_t)\nonumber\\
&\quad 
\leq 
 \frac{\gamma}{2}\left\|\hbx_t-\bbx_t\right\|^2
+(\hbx_t-\bbx_t)^T\nabla f_t(\bbx_t),
\end{align}
which contradicts the result in \eqref{proof_update_3}. Therefore, the assumption that $\hbx_t\neq\tbx_t$ leads to a contradiction. Hence, we can conclude that $\hbx_t=\tbx_t$ which implies that the updates in \eqref{auxiliary_update} and \eqref{eqi_update} are equivalent.

\subsection{Proof of Proposition \ref{prop_lin}}\label{prop_lin_app}

Strong convexity of the function $f_t$ implies that
\begin{equation}\label{proof_10}
f_t(\bbx)- \frac{\mu}{2} \|\bbx-\bbx_t\|^2\geq f_t(\bbx_t) +\nabla f_t(\bbx_t)^T(\bbx-\bbx_t) ,
\end{equation}
for any $\bbx\in \cal{X}$. By adding and subtracting the inner product $\nabla f_t(\bbx_t)^T(\hbx_t-\bbx_t) $ to the right hand side of \eqref{proof_10} we obtain 
\begin{align}\label{proof_20}
&f_t(\bbx)- \frac{\mu}{2} \|\bbx-\bbx_t\|^2\\
&\quad \geq f_t(\bbx_t) +\nabla f_t(\bbx_t)^T(\hbx_t-\bbx_t)+\nabla f_t(\bbx_t)^T(\bbx-\hbx_t) .\nonumber
\end{align}
Observe that the optimality condition of the update of $\hbx_t$ in \eqref{eqi_update} implies that 
\begin{equation}\label{proof_30}
\left(\nabla f_{t}(\bbx_t)+\gamma(\hbx_t-\bbx_t)\right)^T(\bbx-\hbx_t)\geq0,
\end{equation}
for any $\bbx \in \cal{X}$. From the result in \eqref{proof_30}, it follows that the inner product $\nabla f_{t}(\bbx_t)^T(\bbx-\hbx_t)$ is bounded below by $\gamma (\bbx_t-\hbx_t)^T(\bbx-\hbx_t)$. Applying this substitution into \eqref{proof_20} yields 
\begin{align}\label{proof_40}
&f_t(\bbx)- \frac{\mu}{2} \|\bbx-\bbx_t\|^2\\
&\quad \geq f_t(\bbx_t) +\nabla f_t(\bbx_t)^T(\hbx_t-\bbx_t)+\gamma (\bbx_t-\hbx_t)^T(\bbx-\hbx_t) .\nonumber
\end{align}
According to the Lipschitz continuity of the instantaneous gradients $\nabla f_t$ in Assumption \ref{Lipschitz_gradients_ass} and Taylor's series of the objective function $f_t(\hbx_t)$ near the point $\bbx_t$ we can write 
\begin{align}\label{proof_50}
f_t(\hbx_t)&\leq f_t(\bbx_t)\! +\!\nabla f_t(\bbx_t)^T\!(\hbx_t-\bbx_t)+\frac{L}{2}\|\hbx_t\!-\!\bbx_t\|^2\nonumber\\
&\leq f_t(\bbx_t) \!+\!\nabla f_t(\bbx_t)^T\!(\hbx_t-\bbx_t)+\frac{\gamma}{2}\|\hbx_t\!-\!\bbx_t\|^2,
\end{align}
where the second inequality holds since $\gamma\geq L$. Thus, the sum $f_t(\bbx_t) +\nabla f_t(\bbx_t)^T(\hbx_t-\bbx_t)$ is bounded below by $f_t(\hbx_t)- (\gamma/2)\|\hbx_t-\bbx_t\|^2$. By applying this substitution into \eqref{proof_40} we obtain
\begin{align}\label{proof_60}
&f_t(\bbx)- \frac{\mu}{2} \|\bbx-\bbx_t\|^2\nonumber\\
&\quad \geq f_t(\hbx_t) -\frac{\gamma}{2}\|\hbx_t-\bbx_t\|^2+\gamma (\bbx_t-\hbx_t)^T(\bbx-\hbx_t) .
\end{align}
By adding and subtracting $\bbx_t$ we can expand the inner product $(\bbx_t-\hbx_t)^T(\bbx-\hbx_t) $ as the sum of $(\bbx_t-\hbx_t)^T(\bbx-\bbx_t) $ and $(\bbx_t-\hbx_t)^T(\bbx_t-\hbx_t) $. From applying this substitution into \eqref{proof_60} it follows that
\begin{align}\label{proof_80}
&f_t(\bbx)- \frac{\mu}{2} \|\bbx-\bbx_t\|^2
\nonumber\\
&\quad \geq f_t(\hbx_t) +\frac{\gamma}{2}\|\hbx_t-\bbx_t\|^2+\gamma (\bbx_t-\hbx_t)^T(\bbx-\bbx_t) .
\end{align}
Now set $\bbx=\bbx_t^*$ in \eqref{proof_80} and regroup the terms to obtain
\begin{align}\label{proof_81}
&f_t(\bbx_t^*)-  f_t(\hbx_t) 
\\
&\ \geq\frac{\mu}{2} \|\bbx_t^*-\bbx_t\|^2+\frac{\gamma}{2}\|\hbx_t-\bbx_t\|^2+\gamma (\bbx_t-\hbx_t)^T(\bbx_t^*-\bbx_t) .\nonumber
\end{align}
Note that the optimal objective function value $f_t(\bbx_t^*)$ is smaller than $f_t(\hbx_t)$. Thus, the left hand side of \eqref{proof_81} is non-positive which implies that the right hand side is also smaller than $0$, i.e., $({\mu}/{2}) \|\bbx_t^*-\bbx_t\|^2+({\gamma}/{2})\|\hbx_t-\bbx_t\|^2+\gamma (\bbx_t-\hbx_t)^T(\bbx_t^*-\bbx_t)\leq0$. Therefore, by dividing both sides of the inequality by $\gamma$ and regrouping the terms it follows that 
\begin{equation}\label{proof_90}
(\bbx_t-\hbx_t)^T(\bbx_t-\bbx_t^*) \geq\frac{1}{2}\|\hbx_t-\bbx_t\|^2+ \frac{\mu}{2\gamma} \|\bbx_t^*-\bbx_t\|^2  .
\end{equation}

Now by showing a lower bound for the inner product $(\bbx_t-\hbx_t)^T(\bbx_t-\bbx_t^*) $ in terms of the squared norms $\|\hbx_t-\bbx_t\|^2$ and $\|\bbx_t^*-\bbx_t\|^2$ we proceed to prove the claim in \eqref{lin_convg_claim}. Consider the update $ \bbx_{t+1} = (1-h)\bbx_t + h \hbx_t $ in \eqref{variable_update}. By subtracting $\bbx_t^*$ from both sides of the equality and computing the squared norm of the resulted expression we obtain
\begin{align}\label{proof_100}
&\|\bbx_{t+1}-\bbx_t^*\|^2\\
&= \|\bbx_{t}-\bbx_t^*\|^2+h^2\|\bbx_t-\hbx_t\|^2-2h(\bbx_{t}-\bbx_t^*)^T(\bbx_t-\hbx_t).\nonumber
\end{align}
Substitute the inner product $(\bbx_{t}-\bbx_t^*)^T(\bbx_t-\hbx_t)$ in \eqref{proof_100} by its lower bound in \eqref{proof_90} to obtain
\begin{align}\label{proof_110}
&\|\bbx_{t+1}-\bbx_t^*\|^2\nonumber\\
&\leq\left(1-\frac{h\mu}{\gamma}\right)\|\bbx_{t}-\bbx_t^*\|^2+ h(h-1)\|\bbx_t-\hbx_t\|^2.
\end{align}
First note that $h$ is a constant from the interval $(0,1]$, therefore, the term $h(h-1)\|\bbx_t-\hbx_t\|^2$ is smaller than zero and we can simplify the right hand side of \eqref{proof_110} as
\begin{align}\label{proof_120}
\|\bbx_{t+1}-\bbx_t^*\|^2
\leq\left( 1-\frac{h\mu}{\gamma}\right) \|\bbx_{t}-\bbx_t^*\|^2.
\end{align}
Note that the strong convexity constant $\mu$ is smaller than the constant of gradients Lipschitz continuity $L$, i.e., $\mu\leq L$. Thus, we obtain that $\mu\leq \gamma$, since the constant $\gamma$ is chosen such that $\gamma\geq L.$ Moreover, $h\leq1$ which implies that $0<h\mu/\gamma\leq1$. Thus, the coefficient in in \eqref{proof_120} satisfies $0\leq 1- h\mu/\gamma<1$. Hence, the constant $\rho:=(1- h\mu/\gamma)^{1/2}$ is from the interval $[0,1)$. This observation in conjunction with the inequality in \eqref{proof_120} follows the claim in \eqref{lin_convg_claim}.

\subsection{Proof of Theorem \ref{thm_OGD}}\label{thm_OGD_app}

We use the result in Proposition \ref{prop_lin}  to derive an upper bound for the aggregate error $\sum_{t=1}^T \|\bbx_{t}-\bbx_t^*\| $. First, decompose the sum $\sum_{t=1}^T \|\bbx_{t}-\bbx_t^*\| $ as the sum of the first term $\|\bbx_{1}-\bbx_1^*\|  $ and the remaining terms $\sum_{t=2}^T \|\bbx_{t}-\bbx_t^*\| $. Now using the triangle inequality we can show that each term $\|\bbx_t-\bbx_t^*\|$  for $t=2,\dots ,T$ is bounded above by the sum $\|\bbx_{t}-\bbx_{t-1}^*\| + \|\bbx_{t}^*-\bbx_{t-1}^*\| $. Considering this upper bound we can show that the aggregate error $\sum_{t=1}^T \|\bbx_{t}-\bbx_t^*\|  $ is bounded above by
\begin{align}\label{lin_proof_10}
&\sum_{t=1}^T \|\bbx_{t}-\bbx_t^*\| \\
&\quad \leq \|\bbx_{1}-\bbx_1^*\| +  \sum_{t=2}^T \|\bbx_{t}-\bbx_{t-1}^*\| + \sum_{t=2}^T \|\bbx_{t}^*-\bbx_{t-1}^*\|  \nonumber.
\end{align}
According to the result in \eqref{lin_convg_claim}, the sum $\sum_{t=2}^T \|\bbx_{t}-\bbx_{t-1}^*\| $ is bounded above by $\rho\sum_{t=2}^T \|\bbx_{t-1}-\bbx_{t-1}^*\| $. Replace the sum $\sum_{t=2}^T \|\bbx_{t}-\bbx_{t-1}^*\| $ in \eqref{lin_proof_10} by the upper bound $\rho\sum_{t=2}^T \|\bbx_{t-1}-\bbx_{t-1}^*\| $, and add and subtract the term $\rho \| \bbx_T-\bbx_T^*\|$ to the right hand side of the resulted expression   to obtain 
\begin{align}\label{lin_proof_30}
\sum_{t=1}^T \|\bbx_{t}-\bbx_t^*\|   &\leq \|\bbx_{1}-\bbx_1^*\| -\rho \| \bbx_T-\bbx_T^*\| \\
& \quad +\rho  \sum_{t=1}^T \|\bbx_{t}-\bbx_{t}^*\|+ \sum_{t=2}^T \|\bbx_{t}^*-\bbx_{t-1}^*\|. \nonumber
\end{align}
Regrouping the term in \eqref{lin_proof_30} implies that the aggregate error $ \sum_{t=1}^T \|\bbx_{t}-\bbx_t^*\| $ is bounded above by
\begin{align}\label{lin_proof_40}
\sum_{t=1}^T \|\bbx_{t}-\bbx_t^*\| &\leq  \frac{\|\bbx_{1}-\bbx_1^*\| -\rho \| \bbx_T-\bbx_T^*\| }{(1-\rho)} \nonumber\\
&\qquad + \frac{1}{(1-\rho)}\sum_{t=2}^T \|\bbx_{t}^*-\bbx_{t-1}^*\| .
\end{align}
Considering the definitions of $K_1$ and $K_2$ in \eqref{C_defs}, and the result in \eqref{lin_proof_40} it follows that the claim in \eqref{lin_convg_claim} holds and the proof is complete.